\documentstyle[aaai]{article}

\newtheorem{df}{Definition}
\newtheorem{prop}{Proposition}
\newtheorem{thm}{Theorem}

\newtheorem{cor}{Corollary}

\newcommand{\ca}{\mid\!\sim}
\newcommand{\nca}{\not\kern-0.07cm\mid\!\sim}
\newcommand{\uu}{{\mathcal U}}
\newcommand{\vv}{{\mathcal V}}
\newcommand{\ww}{{\mathcal W}}
\newcommand{\sent}{\textrm{sent}}

\newcommand{\rank}{\textrm{rank}}

\title{The lexicographic closure as a revision process} 

\author{Richard Booth \\
	University of Leipzig \\
	Dept.\ of Computer Science \\
	Augustusplatz 10/11 \\
	04109 Leipzig, Germany \\
	booth@informatik.uni-leipzig.de}

\begin{document}

\maketitle

\begin{abstract}
\begin{quote}
	The connections between nonmonotonic reasoning and belief revision
	are well-known. 
	A central problem in the area of nonmonotonic reasoning is the
	problem of default entailment, i.e., when should an item of default 
	information representing ``if $\theta$
	is true then, normally, $\phi$ is true'' be said to follow from
	a given set of items of such information. Many answers to this
	question have been
	proposed but, surprisingly, virtually none have attempted any explicit
 	connection to belief revision. The aim of this paper is to give an
	example of how such a connection can be made by showing how the 
	lexicographic closure of a set of defaults may be conceptualised as 
	a process
	of iterated revision by sets of sentences. Specifically we use
	the revision method of Nayak.
\end{quote}
\end{abstract}

\section{Introduction and Preliminaries}

	The methodological connections between the areas of nonmonotonic 
	reasoning, i.e., the 
	process by which an agent may, possibly, withdraw previously derived
	conclusions upon enlarging her set of hypotheses 
	\cite{mak}, and belief revision,
	i.e., the process by which an agent changes her beliefs upon 
	discovering some new information
	\cite{agm,gardenf}, 
	are well-known
	(see, for example, \cite{garmak,garott,makgar,rott3}). As a 
	consequence, it is possible to translate 
	particular problems in one area into problems in the other. One
	particular problem in nonmonotonic reasoning is the question of
	{\em default entailment}, i.e., when should we regard one item of 
	so-called
	``default knowledge'' (hereafter just ``default''),
	i.e., an expression of the form $\theta \Rightarrow \phi$ standing 
	for ``if $\theta$ then normally (or
	usually, or typically) $\phi$'',
	as ``following from'' a given set of defaults. Several answers to
	this question have been proposed in the literature (such as in
	\cite{benfer,bss,gmp,lehm,lehmag,pearl,emil1}, to name but a few) but 
	none of them (with
	the exception of the last named)
	seem to attempt any explicit connection with belief revision. 
	The aim of this paper is to make a start on such a connection by
	showing how one particular method of default entailment, namely
	the lexicographic closure construction \cite{benfer,lehm}
	can be given a formulation in terms of a certain method of belief
	revision which was first given in \cite{nayak2} and studied further
	in \cite{nayak}. In 
	the process, we uncover one or two
	interesting avenues for further research on both sides.  
	
	The plan of this paper is as follows. Firstly, in the next section 
	we formally pose the basic 
	question of default entailment outlined above and describe
	the lexicographic closure. The set of defaults 
	defined by the lexicographic closure, considered as a binary relation,
 	forms a rational consequence relation (in the sense of \cite{klm}).
	The section following this 
	introduces the theory of belief revision and the important notion of 
	epistemic 
	entrenchment relation (E-relation for short) which it utilises. Also 
	in this section we describe 
	the correspondence between E-relations and rational consequence
	relations. Next, with the aid of this correspondence, 
	we describe Nayak's operation of 
	revision. Nayak proposes
	to model revision of an epistemic state (represented as an  
	E-relation) by an arbitrary set of sentences by first converting 
	this set into
	an E-relation and then revising by this relation. 
	We present one particular method for generating an E-relation
	from a set of sentences and show our main result: that, given this 
	method, the  
	E-relation corresponding to the lexicographic closure can be obtained
	by revising the initial epistemic state (which we take to be the 
	E-relation in
	which the only sentences believed are the tautologies) firstly by
	the set of (the material counterparts of) those defaults which are
	the least specific, then those defaults which are the 
	next-least
	specific and so on up to the set of the most specific defaults.
	After this we give our ideas for possible further study
	before offering some short concluding remarks.

	Before we get started, let us fix our notation.
	Throughout this paper, $L$ is an arbitrary but fixed 
	propositional language built up from a finite set of propositional
	variables using the usual connectives
	$\neg, \wedge, \vee, \rightarrow, \top$ and $\bot$. Semantics is 
	provided by the (finite) set $W$ of propositional
	worlds. For $\theta \in L$ 
	we set $S_\theta
	= \{w \in W \mid w \models \theta\}$, i.e., $S_\theta$ is the set of
	worlds which satisfy $\theta$. Given 
	$E \cup \{\phi\} \subseteq L$ we write $E \models \phi$ 
	whenever $\bigcap_{\theta \in E}S_\theta
	\subseteq S_\phi$ and let $Cn(E)$ denote the set $\{\phi \mid
	E \models \phi\}$. As usual we write $\theta
	\models \phi$ rather than $\{\theta\} \models \phi$ etc.\ while,
	for any $w \in W$
	and $E \subseteq L$ we set $\sent_E(w) = \{\theta \in E \mid
	w \models \theta\}$. Finally, for an arbitrary set $X$ we use $|X|$ 
	to denote the cardinality of $X$.

\section{The Lexicographic Closure of a Set of Defaults}

	Suppose we have somehow learnt
	that an intelligent agent believes some finite set of 
	defaults \mbox{$\Delta
	= \{\lambda_i
	\Rightarrow \chi_i \mid \lambda_i, \chi_i \in L, i = 1, \ldots, l\}$.}
	In this case what other assertions of this form
	should we conclude our agent believes? Or, put another way, what is 
	the binary relation
	$\ca^\Delta$ on $L$ where $\theta \ca^\Delta \phi$ holds iff we can 
	conclude, on the basis of $\Delta$, that if $\theta$ is true then,
	normally, $\phi$ is also true? In this paper, one answer to
	this question which we are 
	particularly interested in is 
	the lexicographic closure construction 
	which was proposed 
	independently in both \cite{benfer} and 
	\cite{lehm}. We describe this construction now.
	
	Throughout this paper we assume that $\Delta$ is an arbitrary 
	but fixed,
	finite set of defaults. For this paper we also make the 
	simplifying 
	assumption that $\Delta$ is ``consistent'', in the sense that its
	set of {\em material counterparts} $\Delta^\rightarrow =
	\{\lambda \rightarrow \chi \mid \lambda 
	\Rightarrow 
	\chi \in \Delta\}$ is consistent.
	Using a procedure
	given in \cite{pearl} (or, equivalently, in \cite{lehmag}) we may 
	partition $\Delta$
	into $\Delta = (\Delta_0, \ldots, \Delta_n)$, where the $\Delta_i$
	correspond, in a precise sense, to ``levels of specificity'' --
	given a default $\delta \in \Delta$, the larger the $i$ for which
	$\delta \in \Delta_i$, the more specific are the situations
	to which $\delta$ is applicable. Following \cite{pearl}, we call this 
	partition the {\em Z-partition} of $\Delta$. Like many methods of
	default entailment (see \cite{benfer} for several examples), the 
	lexicographic closure can 
	be based on a 
	method of choosing maximal consistent subsets of $\Delta^\rightarrow$. 
	More 
	precisely the lexicographic closure is a member of a 
	family of 
	consequence relations $\ca^\Delta_\ll$, where $\ll$ is an
	ordering on $2^\Delta$, and, for all $\theta, 
	\phi \in L$, we have
\[
\begin{array}{rcl}
	\theta \ca^\Delta_{\ll} \phi
	& \textrm{iff} &
	\textrm{for all}\ \Gamma \subseteq \Delta\ \textrm{such that}\
	\Gamma^\rightarrow \cup \{\theta\}\ \textrm{is} \\
	& &
	\textrm{consistent and}\
	\Gamma\ \textrm{is $\ll$-maximal amongst}	\\
	& &
	\textrm{such subsets, we have}\
	\Gamma^\rightarrow \cup \{\theta\} \models \phi.
\end{array}
\]
	To specify the lexicographic closure we 
	instantiate the order $\ll$ above, with the help of the Z-partition, 
	as follows:
	Given subsets $A, B \subseteq \Delta$
	let $A_i = A \cap \Delta_i$ and $B_i = B \cap \Delta_i$ for
	each $i = 0, \ldots, n$. 
	We define an ordering $\ll_{lex}$ on $2^\Delta$ by:
\[
\begin{array}{rcl}
	A \ll_{lex} B &
	\textrm{iff} &
	\textrm{there exists $i$ such that}\ |A_i| < |B_i|\
	\textrm{and,}						\\
	& & \textrm{for all}\ j > i,\ |A_j| = |B_j|.	
\end{array}
\] 
	(The reason for the name ``{\em lexicographic} closure'' should now be
	clear.) The lexicographic closure $\ca^\Delta_{lex}$ is then just 
	defined to be $\ca^\Delta_{\ll_{lex}}$.

	How successful is $\ca^\Delta_{lex}$ in achieving the goals
	of default reasoning? We refer the reader to \cite{lehm} for the
	details. However, the internal, closure properties
	of $\ca^\Delta_{lex}$ can be summed up by the following proposition,
	which can be found jointly in \cite{benfer} and \cite{lehm}. 

\begin{prop}
\label{lexrat}
	The binary relation $\ca^\Delta_{lex}$ is a 
	rational consequence relation (see \cite{klm,lehmag}).
	Furthermore, $\ca^\Delta_{lex}$ is consistency
	preserving, i.e., for all $\theta$, \mbox{$\theta \ca^\Delta_{lex} 
	\bot$} implies $\theta \models \bot$.
\end{prop}
	Now we already know (see, for
	example, \cite{freund2,lehmag}) that rational 
	consequence relations may be represented by finite sequences
	$\vec\uu = (\uu_0, \ldots, \uu_k)$ of mutually disjoint subsets of $W$
 	in the following 
	sense: Given such a sequence $\vec\uu$ and $\theta \in L$ we set
	$\rank^{\vec\uu}(\theta) =$ the least $i$ such that $\uu_i \cap S_\theta \neq
	\emptyset$. If no such $i$ exists then we set $\rank^{\vec\uu}(\theta) = \infty$.
	If we then define a binary relation $\ca_{\vec\uu}$ on $L$ by 
	setting\footnote{Note the first clause includes the 
case $\rank^{\vec\uu}(\theta \wedge
	\neg\phi) = \infty$ and $\rank^{\vec\uu}(\theta) \neq \infty$.}
\[ 
\begin{array}{rccl}
	\theta \ca_{\vec\uu} \phi
	& \textrm{iff} & \textrm{either} &
	\rank^{\vec\uu}(\theta) < \rank^{\vec\uu}(\theta \wedge \neg\phi) \\
	& & \textrm{or}& 
	\rank^{\vec\uu}(\theta) = \infty
\end{array}
\]
	then $\ca_{\vec\uu}$ forms a rational consequence relation, 
	while
	moreover every rational consequence relation arises in this way
	from some sequence $\vec\uu$.\footnote{Such sequences are clearly 
equivalent to the ranked models used to characterise rational 
	consequence relations in \cite{lehmag}.}
 	The intuition behind the
	sequences $\vec\uu$ is that they represent a ``ranking'' of the worlds
	in $W$ according to their plausibility -- the lower the $i$ for which
	$w \in \uu_i$, the more plausible, in relation 
	to the other worlds,
	it is considered to be. If $w \not\in \uu_i$ for all $i$ then
	we may take $w$ to be considered ``impossible''. 

	One thing to note about the definition of $\ca_{\vec\uu}$ given above
	is that we allow $\emptyset$
	to appear, possibly more than once, in $\vec\uu$.\footnote{This 
approach carries us very close to the ``semi-quantitative'' approaches of 
\cite{spohn,emil1,williams}, which use an explicit ranking 
function as a starting
point rather than deriving one from a sequence of world-sets. Our
approach, though, remains squarely qualitative in character.}
	This freedom comes
	in useful when proving some of our results. It also has the effect
	that the mapping $\vec\uu \mapsto \ca_{\vec\uu}$ detailed above is not 
	injective -- given a rational consequence relation $\ca$ there will
	be many (in fact infinitely many) sequences $\vec\uu$ such that
	$\ca = \ca_{\vec\uu}$.\footnote{Since clearly we can insert as many 
copies of $\emptyset$ into the sequence $(\uu_0, \ldots, \uu_k)$ as we wish 
without changing the relation $\ca_{\vec\uu}$.} 
	Another thing to note
	about $\ca_{\vec\uu}$ is that $\ca_{\vec\uu}$ 
	will be consistency preserving iff
	$\bigcup_{i=0}^k \uu_i = W$, while it will be trivial, i.e., will 
	satisfy $\theta \ca_{\vec\uu} \phi$ for {\em all} $\theta$ and $\phi$,
	iff
	$\bigcup_{i=0}^k \uu_i = \emptyset$. We make the following 
	definitions:
\begin{df}
	Let $\vec\uu = (\uu_0, \dots, \uu_k)$ be a finite sequence of 
	mutually disjoint subsets
	of $W$. We shall say that $\vec\uu$ is {\em full} iff 
	$\bigcup_{i=0}^k \uu_i = W$ and that $\vec\uu$ is {\em empty} iff
	$\bigcup_{i=0}^k \uu_i = \emptyset$. We let $\Upsilon$ denote the
	set of all such $\vec\uu$ which are either full or empty.
\end{df}
	Hence Proposition \ref{lexrat} tells us that there must exist a 
	full sequence $\vec\uu \in \Upsilon$ such 
	that $\theta \ca^\Delta_{lex} \phi$ iff $\theta \ca_{\vec\uu} 
	\phi$. What form does $\vec\uu$ take here? The answer is given in 
	\cite{benfer} and
	\cite{lehm} (and is, in fact, used to {\em define} $\ca^\Delta_{lex}$
	in the latter). In this paper we show that we can 
	arrive at this answer via a different route.

\section{Belief Revision and Epistemic Entrenchment}

	Belief revision is concerned with
	the following problem: How should an agent revise her beliefs upon
	receiving some new information which may, possibly, contradict some
	of her current beliefs? The most popular basic framework within which 
	this
	question is studied is the one laid down by Alchour\'ron, G\"ardenfors
	and Makinson (AGM) in \cite{agm}. In that 
	framework an agent's epistemic state is represented as a logically
	closed set of sentences called a {\em belief set}, and the new 
	information, or epistemic input, is represented as a single sentence.
	AGM propose a number of postulates which a reasonable operation
	of revision should satisfy. In particular, the revised 
	belief set
	should contain the epistemic input and 
	should be consistent.\footnote{Unless 
the epistemic input itself is inconsistent. See \cite{gardenf}
for the full list of postulates with detailed discussion.} 
	In order to meet these requirements, in the general case when the 
	input is inconsistent with the prior belief set, the agent is
	forced to give up some of her prior beliefs. One way of determining
	precisely {\em which} sentences the agent should give up in this 
	situation is to assign to the agent an
	E-relation $\preceq$ on $L$
	(see, for example, \cite{gardenf,garmak,nayak2,rott,rott3}). 

	The intuitive meaning behind E-relations is that
	\mbox{$\phi \preceq \psi$} should hold iff the agent finds it at 
	least as
	easy to give up $\phi$ as she does $\psi$, i.e., her belief in $\psi$
	is at least as entrenched as her belief in $\phi$. In cases of 
	conflict the
	agent should then give up those sentences which are less
	entrenched.  
	In what follows we use $\prec$ to denote the strict part of $\preceq$,
	i.e, $\theta \prec \phi$ iff $\theta \preceq \phi$ and not($\phi 
	\preceq \theta$).
	We follow \cite{nayak2} in formally defining 
	E-relations as follows:
\begin{df}
	An {\em epistemic entrenchment relation} (E-relation) (on $L$) is a 
	relation
	$\preceq \subseteq L \times L$ which satisfies the following 
	conditions for all $\theta, \phi, \psi \in L$, \\
\begin{tabular}{rl}
	(E1) & If $\theta \preceq \phi$ and $\phi \preceq \psi$ then 
	$\theta \preceq \psi$     			\\
	& 
	\hfill (transitivity)			\\
	(E2) & If $\theta \models \phi$ then $\theta \preceq \phi$ 
	\hfill (dominance)						\\
	(E3) & $\theta \preceq \theta \wedge \phi$ or $\phi \preceq \theta 
	\wedge \phi$ \hfill (conjunctiveness)				\\
	(E4) & Given there exists $\psi \in L$ such that $\bot \prec
	\psi$, \\
	& if $\theta \preceq \phi$ for all $\theta \in L$, then $\models
	\phi$  \\
	& \hfill (maximality) 
\end{tabular}
\end{df}
	If there is no $\psi \in L$ such that $\bot \prec \psi$, equivalently,
	if $\theta \preceq \phi$ holds for {\em all} $\theta, \phi$, then
	we call $\preceq$ the {\em absurd} E-relation.
	The original definition of E-relation, such as is found in 
	\cite{gardenf},
	is given relative to a belief set. However, as is noted in 
	\cite{nayak2}, E-relations contain enough information by themselves
	for the belief set to be extracted from it. 
	The belief set $Bel(\preceq)$ associated with the E-relation $\preceq$
	is defined as:
\[
	Bel(\preceq)
	=
	\left\{
	\begin{array}{ll}
		\{\theta \mid \bot \prec \theta\} &
		\textrm{if}\ \bot \prec \theta,\ \textrm{for some}\ \theta,
								\\
		L &
		\textrm{otherwise.}
	\end{array}
	\right.
\]
	The belief set
	associated with an E-relation was called its {\em epistemic content}
	in \cite{nayak2}. 

\subsection{E-relations and Rational Consequence}
	We now bring in the connection 
	between E-relations, as they have been defined here, and rational 
	consequence relations. The following result is virtually the same
	as one given in \cite{garmak}.

\begin{prop}
\label{central}
	Let $\ca$ be a rational consequence relation which is either
	consistency preserving or trivial. If we define, from
	$\ca$, a binary relation $\preceq_\sim$ on $L$ by
	setting, for all $\theta, \phi \in L$,
\begin{equation}
\label{rattoee}
	\theta \preceq_\sim \phi\
	\textrm{iff}\
	\neg\theta \vee \neg\phi \nca \theta\
	\textrm{or}\	
	\neg\phi \ca \bot,
\end{equation} 
	then $\preceq_\sim$ forms an E-relation. Conversely if, given
	an  
	E-relation $\preceq$ we define a binary relation $\ca_\preceq$ on $L$
	by setting, for all $\theta, \phi \in L$,
\[
	\theta \ca_\preceq \phi\
	\textrm{iff}\
	\neg\theta \prec \neg\theta \vee \phi\
	\textrm{or}\	
	\top \preceq \neg\theta
\]
	then $\ca_\preceq$ forms a rational consequence relation
	which is either consistency preserving or trivial. Furthermore
	the identity $\ca = \ca_{\preceq_\sim}$ holds.
\end{prop}
	So there is a bijection between rational consequence relations which
	are either consistency preserving or trivial, and E-relations. 
	Essentially they are different ways of describing the same thing,
	and so an operation for changing one automatically 
	gives us an operation for changing the other. This observation is at
	the heart of the present paper.
	Given $\vec\uu \in \Upsilon$ we shall denote by 
	$\preceq_{\vec\uu}$ the E-relation defined from $\ca_{\vec\uu}$
	via (\ref{rattoee}) above. Since we have already seen 
	that rational consequence relations which
	are either consistency preserving or trivial are characterised
	by the sequences in $\Upsilon$, Proposition \ref{central} leads us to
	the following result.
\begin{prop}
\label{eechar}
	Let $\preceq$ be a binary relation on $L$. Then $\preceq$ is an
	E-relation iff $\preceq = \preceq_{\vec\uu}$ for some $\vec\uu \in
	\Upsilon$.
\end{prop}
	Note again that $\preceq_{\vec\uu} = \preceq_{\vec\vv}$ does not
	imply $\vec\uu = \vec\vv$. Also note that $\preceq_{\vec\uu}$ will be 
	absurd iff $\vec\uu$ is empty. It is straightforward to prove the
	following.
\begin{prop}
\label{start}
	Let $\vec\uu \in \Upsilon$ and $\theta \in L$. Then \mbox{$\theta \in
	Bel(\preceq_{\vec\uu})$} iff $\top \ca_{\vec\uu} \theta$.
\end{prop}

\section{Revision of E-relations}

	Nayak \cite{nayak2} deviates from the basic 
	AGM framework in two ways. Firstly,
	in order to help us deal with {\em iterated} revision
	(see \cite{boutilier,dp,williams}),
	he argues that we need not only a description of the new belief 
	set
	which results from a revision, but also a new E-relation which
	can then guide any further revision. Thus we 
	should
	enlarge our epistemic state to consist of a belief set {\bf together
	with} an E-relation and then perform revision on this larger state. 
	In fact, since, as we have seen, the belief set may be 
	determined
	from the E-relation, we may take our epistemic states to be just
	E-relations.\footnote{In this context of iterated revision, the consideration of more 
comprehensive
epistemic states of which a belief set is but one component has also been
suggested in \cite{dp} and \cite{fh}.} 
	Secondly, he suggests that the epistemic input should consist not
	of a single sentence, but rather another E-relation. (See 
	\cite{nayak2} for motivation.)
	He claims it is then possible, in his framework, to capture
	the revision of E-relations by arbitrary sets of sentences $E$ by 
	first
	converting the set $E$ into a suitable E-relation $\preceq_E$ 
	and then 
	revising by $\preceq_E$. We shall discuss this point further in the 
	next section. In this section we shall use the characterisation
	of E-relations given in Proposition \ref{eechar} to describe
	Nayak's proposal of how one E-relation should be revised by another
	to obtain a new E-relation. The ideas behind this formulation can
	also be seen in \cite{nayak2}.
 
	Let $\preceq_K$ be the prior E-relation and let $\preceq_E$ be
	the input E-relation.
	By Proposition \ref{eechar}, we know that there exist $\vec\uu, 
	\vec\vv \in \Upsilon$
	such that $\preceq_K = \preceq_{\vec\uu}$ and $\preceq_E = 
	\preceq_{\vec\vv}$. Hence we may reduce the question of entrenchment 
	revision to a question of how to revise one sequence of world-sets
	by another. More precisely, we can define a {\em sequence}
	revision function $*: 
	\Upsilon \times
	\Upsilon \rightarrow \Upsilon$, where \mbox{$\vec\uu*\vec\vv$} is the 
	result
	of revising $\vec\uu$ by $\vec\vv$, and then simply lift this to
	an {\em entrenchment} revision function by setting
\begin{equation}
\label{our}
	\preceq_K * \preceq_E
	=
	\preceq_{\vec\uu * \vec\vv}.
\end{equation}
	(The context will always make it clear whether we are considering
	$*$ as an operation on sequences or an operation on E-relations.)
	All this must be independent of precisely
	which $\vec\uu$ and $\vec\vv$ are chosen to represent $\preceq_K$
	and $\preceq_E$ respectively. 
	The definition for the sequence revision function $*$ which we choose,
	motivated purely in order to arrive at Nayak's entrenchment revision
	function, is the following:
\begin{df}
\label{rev}
	We define the function $*:\Upsilon \times \Upsilon \rightarrow 
	\Upsilon$ by setting , for all $\vec\uu = (\uu_0, \ldots, 
	\uu_k)$ and $\vec\vv = (\vv_0, \ldots,
	\vv_m)$, 
\[
	\vec\uu * \vec\vv =
	\left\{
	\begin{array}{l}
		\begin{array}{l}
			(\uu_0 \cap \vv_0, \uu_1 \cap \vv_0, \ldots, \uu_k 
		\cap \vv_0,
									\\
	\uu_0 \cap \vv_1, \uu_1 \cap \vv_1, \ldots, \uu_k \cap \vv_1,
									\\
	\ldots,								\\
	\uu_0 \cap \vv_m, \uu_1 \cap \vv_m, \ldots, \uu_k \cap \vv_m).
		\end{array}  
		\hfill \! \! \textrm{if}\ \vec\uu\ \textrm{is full}
	\vspace{0.3cm}		\\
		\vec\vv \hfill \textrm{otherwise.}
	\end{array}
	\right.
\]
\end{df}
	Clearly it is the case that $\vec\uu * \vec\vv$ is always full,
	unless $\vec\vv$ is empty, in which case so is $\vec\uu * \vec\vv$.
	Hence we certainly have $\vec\uu * \vec\vv \in \Upsilon$.
	The following proposition
	assures us that $*$, when lifted to an operation on E-relations,
	is well-defined.

\begin{prop}
	Let $\vec\uu_i, \vec\vv_i \in \Upsilon$ for $i = 1, 2$. Then
	$\preceq_{\vec\uu_1} = \preceq_{\vec\uu_2}$ and $\preceq_{\vec\vv_1} =
	\preceq_{\vec\vv_2}$
	implies $\preceq_{\vec\uu_1 * \vec\vv_1} = \preceq_{\vec\uu_2 * 
	\vec\vv_2}$.
\end{prop} 

	From now on we will follow Nayak and use $\preceq_{K*E}$ as an
	abbreviation for $\preceq_K * \preceq_E$.
	The authors of \cite{nayak} propose the following postulates for
	the revision of E-relations: \vspace{0.2cm} \\
\begin{tabular}{ll}
(E$1^*$) &	$\preceq_{K * E}$ is an E-relation. \\
(E$2^*$) &	If $\theta \prec_E \phi$ then $\theta \prec_{K*E} \phi$. \\
(E$3^*$) &	If both $\theta \preceq_E \phi$ and $\phi \preceq_E \theta$ 
		and if, for all $\lambda$, \\
	 &	$\chi$ such that $\theta \wedge \phi \models \chi$ and $\theta
		\prec \chi$, we have \\
	 &	$\lambda \preceq_K \chi$ iff $\lambda
		\preceq_E \chi$, then $\theta \preceq_{K*E} \phi$ iff
		$\theta \preceq_K \phi$.
\end{tabular}

\vspace{0.25cm}
\noindent
	We refer the reader to \cite{nayak} for the justification of these
	postulates.
	Any operation of revision of E-relations which satisfies the above
	three conditions is called a {\em well-behaved} entrenchment revision
	operation in \cite{nayak}, where it is shown that there is, in fact,
	precisely one well-behaved entrenchment revision operation, namely
	the one given in \cite{nayak2}. Thus the above three postulates
	serve to characterise Nayak's revision method. Our revision
	operation, defined by Definition \ref{rev} via (\ref{our}) above, 
	also satisfies 
	(E$1^*$)--(E$3^*$) and hence is semantically equivalent to the
	operation constructed in \cite{nayak2}. 
\begin{thm}
	If we set $\preceq_{K*E} = \preceq_{\vec\uu * \vec\vv}$ where
	$\vec\uu$ ($\vec\vv$) is chosen so that $\preceq_K = \preceq_{\vec\uu}$
	($\preceq_E = \preceq_{\vec\vv}$) then the operator $*$ satisfies
	(E$1^*$), (E$2^*$) and (E$3^*$).  
\end{thm}
	
	One advantage of this particular formulation is that it is relatively 
	easy to
	show properties of the well-behaved entrenchment revision operation
	$*$. For example, the following proposition regarding sequence
	revision is straightforward to prove.
\begin{prop}
\label{assoc}
	Let $\vec\uu, \vec\vv, \vec\ww \in \Upsilon$ and suppose $\vec\vv$ is 
	not empty. Then \linebreak $(\vec\uu * \vec\vv)
	* \vec\ww = \vec\uu * (\vec\vv * \vec\ww)$. 
\end{prop}
	This proposition, in turn, gives us the following interesting 
	associativity property of the induced entrenchment revision operation.
\begin{prop}
\label{assocee}
	Let $\preceq_i$ be an E-relation for $i = 1,2,3$. Then, if $\preceq_2$
	is not absurd, we have $(\preceq_1*\preceq_2)*\preceq_3 
	=$ \mbox{$\preceq_1*(\preceq_2*\preceq_3)$.}
\end{prop}

\section{Generating E-relations from Sets of Sentences}

	As we said in the last section, Nayak proposes that his
	way of revising one E-relation by another allows a way of modelling
	the revision of an E-relation by a set of sentences $E$ by first
	converting, according
	to some suitable method, the set $E$ into an E-relation $\preceq_E$ 
	and then 
	revising by $\preceq_E$. The question of which ``suitable method'' 
	we should use for generating
	$\preceq_E$ is clearly an interesting question in itself. A strong
	feeling is that the relation $\preceq_E$ should adequately convey
	the informational content of $E$, but what does this mean? An obvious
	first
	requirement of $\preceq_E$ would seem to be $Bel(\preceq_E) = Cn(E)$,
	but there are different ways in which this can be achieved.
	The definition which Nayak seems to
	advocate is the following, based on an
	idea in \cite{rott}, and expressed via its strict part.
\[
\begin{array}{rcl}
	\theta \prec_E \phi &
	\textrm{iff} &
	E \not\models \bot, \not\models\theta\ \textrm{and}\
	\textrm{for all}\ E' \subseteq E\ 
	\textrm{such that} 				\\
	& &
	E' \cup \{\neg\phi\}\ \textrm{is consistent, there exists}	\\
	& &
	E'' 
	\subseteq E\ 
	\textrm{such that}\ E' \subset E''\ 
	\textrm{and}\ E'' \cup \{\neg\theta\}				\\ 
	& &
	\textrm{is consistent.}
\end{array}	
\]
	The clause ``$E \not\models \bot$'' in the above merely ensures that 
	if $E$ is inconsistent then $\preceq_E$ is absurd, while the clause
	``$\not\models \theta$'' ensures that tautologies are maximally
	entrenched. The main body of the definition essentially says that 
	$\phi$ 
	should be strictly more entrenched than $\theta$
	iff each $\subseteq$-maximal
	subset of $E$ which fails to imply $\phi$ may be strictly enlarged
	to a subset of $E$ which fails to imply $\theta$.
	The problem with defining $\preceq_E$ in this way is that it will
	fail, in general, to be an E-relation. In particular it will not
	necessarily satisfy (E1).\footnote{It should be noted, however, that 
$\preceq_E$ so 
	defined does still enjoy several interesting properties. In fact it
	belongs to Rott's family of {\em generalized} E-relations 
	\cite{rott2}.} How can we modify/extend it so as to 
	obtain an E-relation? The 
	possibility we choose is to compare the sets which fail to imply 
	$\theta$ and $\phi$ by cardinality rather than 
	inclusion:\footnote{Possibilities in this spirit are also discussed in \cite{benfer} 
(Section 2)
and \cite{lehm} (Section 8). 
See also the closely related Section 5 of \cite{freund}.}
\begin{df}
\label{defE}
	Given a set $E \subseteq L$, define
	a relation $\prec_E \subseteq L \times L$ by, for all
	$\theta, \phi \in L$,
\[
\begin{array}{rcl}
	\theta \prec_E \phi &
	\textrm{iff} &
	E \not\models \bot, \not\models\theta\ \textrm{and}\
	\textrm{for all}\ E' \subseteq E\ 
	\textrm{such that}					\\
	& &
	E' \cup \{\neg\phi\}\ \textrm{is consistent, there exists} \\
	& &
	 E'' \subseteq E\ 
	\textrm{such that}\ |E'| < |E''|\ 
	\textrm{and} \\ 
	& &
	E'' \cup \{\neg\theta\}\
	\textrm{is consistent.}
\end{array}	
\]
\end{df}
	Note that this definition does indeed extend the ``old'' definition
	given above. 
	That $\preceq_E$ defined by Definition \ref{defE} is a genuine
	E-relation will follow once we have found a sequence $\vec\uu \in
	\Upsilon$ such that $\preceq_E = \preceq_{\vec\uu}$. We do this as
	follows. Let us assume for simplicity that $E$ is finite with 
	$|E| = k$. Then, for each $i = 0, 
	\ldots, k$, we set
\[
	\uu^E_i = 
	\left\{
	\begin{array}{ll}
	\{w \in W \mid |\textrm{sent}_E(w)| = k-i\} &
	\textrm{if}\ E \not\models \bot				\\
	\emptyset & \textrm{otherwise.}
	\end{array}
	\right.
\]
	So, in the principal case when $E$ is consistent, 
	$\uu^E_i$ contains those worlds which satisfy 
	precisely
	$k-i$ elements of $E$. Let $\vec\uu^E =
	(\uu^E_0, \ldots, \uu^E_k)$. 
\begin{prop}
\label{1}
	If $E \models \bot$ then
	$\vec\uu^E$ is empty, while if $E \not\models \bot$ then 
	$\vec\uu^E$ is full (and so, either way, $\vec\uu^E \in \Upsilon$). 
	In both cases we have
	$\preceq_E = \preceq_{\vec\uu^E}$. Hence $\preceq_E$ is an E-relation.
\end{prop}
	Note that, with
	this notation, we have $\vec\uu^\emptyset = (W)$. Hence we can think 
	of 
	$\preceq_\emptyset$ as being the initial epistemic state in which
	each world is equally plausible.

	How does $\preceq_E$ portray the informational content of $E$?
	The sequence $\vec\uu^E$ shows us clearly.
	First of all 
	it is easy to see that $\preceq_E$ satisfies the basic requirement
	of $Bel(\preceq_E) = Cn(E)$ (in particular the only sentences 
	believed in $\preceq_\emptyset$ are the tautologies) since the most 
	plausible worlds in $\vec\uu^E$, i.e.,
	the worlds in $\uu_0^E$, are precisely those worlds which satisfy
	every sentence in $E$. The big question is how does $\vec\uu^E$
	classify the worlds which do {\bf not} satisfy every sentence in $E$?
	The answer is that it considers one such world more plausible than
	another iff it satisfies strictly more sentences in $E$. This
	makes the 
	relation $\preceq_E$ dependent on the syntactic form, not just the
	semantic form, of $E$, i.e., we can have $Cn(E_1) = Cn(E_2)$ without
	necessarily having $\preceq_{E_1} = \preceq_{E_2}$.
	One situation where this method 
	might be deemed suitable is if 
	we want to regard the elements of $E$ as items of information coming 
	from different, independent sources.
	
	From now on, for the special case when $E$ is a singleton, we shall 
	write 
	$\preceq_\theta$ rather than $\preceq_{\{\theta\}}$ etc. We have the 
	following partial generalisation of Proposition \ref{start}.  

\begin{prop}
\label{key}
	Let $\vec\uu \in \Upsilon$ be full and let $\theta, \phi \in L$. Then
	$\phi \in Bel(\preceq_{\vec\uu}*\preceq_\theta)$ iff $\theta 
	\ca_{\vec\uu} \phi$.
\end{prop}	
	
	We are now ready to give the sequence $\vec\uu$ such that
	\mbox{$\theta \ca_{\vec\uu} \phi$} iff $\theta \ca^\Delta_{lex} \phi$.
	Let
	$(\Delta_0, \ldots, \Delta_n)$ be the Z-partition of $\Delta$. Then,
	to obtain our special $\vec\uu$ we start at the sequence $(W)$ and 
	then successively revise, using our sequence revision function $*$,
	by $\vec\uu^{\Delta^\rightarrow_i}$ for $i = 0, 1, \ldots, n$. 
	Recalling
	that $(W) = \vec\uu^\emptyset$ we may give our main result. Recall
	that we are assuming $\Delta$ is finite and that $\Delta^\rightarrow$
	is consistent.
\begin{thm}
\label{main}
	Let $\Delta$ be a set of defaults with
	associated Z-partition $(\Delta_0, \ldots, \Delta_n)$. Then, for
	all $\theta, \phi \in L$, we have $\theta \ca^\Delta_{lex} \phi$ iff 
	$\theta \ca_{\vec\uu^\emptyset*\vec\uu^{\Delta^\rightarrow_0} * \cdots
	* \vec\uu^{\Delta^\rightarrow_n}} \phi$.
\end{thm}
	Note that, by Proposition \ref{assoc} and the assumption that 
	$\Delta^\rightarrow$ is consistent, the term 
	$\vec\uu^\emptyset*\vec\uu^{\Delta^\rightarrow_0} * \cdots
	* \vec\uu^{\Delta^\rightarrow_n}$ is independent of the bracketing.
	Similar remarks apply (using Proposition \ref{assocee}) to the next
	result.
	Using Propositions \ref{1} and \ref{key} we may re-express Theorem
	\ref{main} as:
\begin{cor}
	Let $\Delta$ be a set of defaults with
	associated Z-partition $(\Delta_0, \ldots, \Delta_n)$. Then, for
	all $\theta, \phi \in L$, we have $\theta \ca^\Delta_{lex} \phi$ iff 
	$\phi \in Bel(\preceq_\emptyset * \preceq_{\Delta^\rightarrow_0} * 
	\cdots * \preceq_{\Delta^\rightarrow_n}*\preceq_\theta)$.
\end{cor}
	If we go further and actually identify a revision of the form
	$\preceq*\preceq_E$ with $\preceq*E$ then we have the following
	characterisation of the lexicographic closure.
\begin{cor}
\label{fin}
	Let $\Delta$ be a set of defaults with
	associated Z-partition $(\Delta_0, \ldots, \Delta_n)$. Then, for
	all $\theta, \phi \in L$, we have $\theta \ca^\Delta_{lex} \phi$ iff 
	$\phi \in
	Bel(\preceq_\emptyset * \Delta^\rightarrow_0 * \cdots * 
	\Delta^\rightarrow_n * \theta)$.
\end{cor}
	Hence, using {\em this} particular method of revision and {\em this}
	particular way of interpreting revision by a set of sentences, we
	have shown that $\theta \ca^\Delta_{lex} \phi$ iff $\phi$ is believed
	after first successively revising the initial epistemic state by
	the set of sentences $\Delta^\rightarrow_i$ for $i = 0,1, \ldots, n$,
	and then revising by $\theta$.
 
\section{Further Work}

	The developments in the previous sections have raised a couple of
	questions regarding both belief revision {\em and} default entailment.
	Firstly, 
	while there have been several papers published 
	concerned
	with iterated revision by single sentences, and also some concerned
	with revision by sets of sentences,\footnote{Either directly (e.g.\ 
\cite{zhang}) or indirectly, via the study
of {\em contraction} by a set of sentences (e.g.\ \cite{fha}). See 
\cite{gardenf} for a description of contractions and their close relationship 
with revision.} 
	there seems to be
	little in the way of any systematic study of iterated revision 
	{\em by} 
	sets of sentences.\footnote{An exception, in a slightly more
complex framework, is \cite{emil2}.}
	Darwiche and Pearl \cite{dp} provide a postulational approach
	to the question of iterated revision of epistemic states by single 
	sentences. In this
	approach they take the concept of epistemic state to be primitive,
	assuming only that from each such state $\Psi$ we may extract a
	belief set (in the usual AGM sense of the term) $B(\Psi)$ 
	representing the set of sentences accepted in that state. 
	For 
	example Darwiche and Pearl's second postulate may be stated as
\[
	\textrm{If}\ \phi \models \neg\theta\
	\textrm{then}\
	B((\Psi * \theta) * \phi) = B(\Psi * \phi).
\]
	(For the other postulates and their justifications see \cite{dp}.)
	It is not difficult to see that, if we identify epistemic state here
	with E-relation and take $B(\preceq) = Bel(\preceq)$, then the method
	proposed by Nayak, on its restriction to single 
	sentences\footnote{We obviously interpret single sentences here as
singleton sets.}
	satisfies all of Darwiche and Pearl's postulates. However, it
	also satisfies some interesting properties in the general case. For
	example, given an E-relation $\preceq$ and $E_1 \subseteq E_2 
	\subseteq L$ such that $E_2$ is consistent, we have $(\preceq * E_2) * 
	E_1 = (\preceq * E_2-E_1)
	* E_1$. In particular, if $\{\theta, \phi\}$ is consistent, we have 
	$(\preceq * \{\theta, \phi\}) * \phi
	= (\preceq * \theta) * \phi$. (Note this is a stronger statement
	than just
	$Bel((\preceq * \{\theta, \phi\}) * \phi) =
	Bel((\preceq * \theta) * \phi)$.) The question of whether this, or
	any other, property of iterated revision by sets is desirable seems to 
	be a question worth investigating. Another question is: Can we,  
	by modifying the various parameters involved in this revision process,
	model any of the other existing 
	methods of
	default entailment, apart from the lexicographic closure, or even
	construct new ones? 
	For example, given our set of defaults $\Delta$ and its Z-partition
	$(\Delta_0, \ldots, \Delta_n)$, let $\Theta_i = \bigcup_{i \leq j}
	\Delta_j$ 
	for each $i = 1, \ldots, n$. Then, by the above comments, we may 
	rewrite Corollary \ref{fin} as
\[
	\theta \ca^\Delta_{lex} \phi\
	\textrm{iff}\
	\phi \in Bel(\preceq_\emptyset * \Theta^\rightarrow_0 * 
	\cdots * \Theta^\rightarrow_n * \theta).
\]
	We conjecture that if we now replace
	each $\Theta^\rightarrow_i$ in the above by $\bigwedge 
	\Theta^\rightarrow_i$ (i.e.,
	the conjunction, in some order, of the sentences in 
	$\Theta^\rightarrow_i$),
	then we obtain the {\em rational closure} \cite{lehmag} (which
	is semantically equivalent to System Z \cite{pearl}) of $\Delta$,
	instead of the lexicographic closure. This and other variations are 
	the subject 
	of ongoing study. Finally, note that, since we assumed at the outset
	that our language $L$ is based on only finitely many propositional
	variables, and also 
	that $\Delta$ is a finite set of defaults, we have not needed in this
	paper to
	confront the question of revision by {\em infinite} sets of sentences.
	It remains to be seen to what extent the ideas in this paper
	can be extended to cover this more general situation.\footnote{For
	one treatment of this topic, and its relation with nonmonotonic
	inference from infinite sets of premises, see \cite{zhang2}.}

\section{Conclusion}

	In this paper we have taken a particular model of default reasoning 
	-- the lexicographic closure -- and re-cast it in terms of iterated
	belief
	revision by sets of sentences, using the particular, independently
	motivated, revision model
	of Nayak. In the process of doing this, a couple of interesting
	avenues for further exploration have suggested themselves. In 
	particular, the questions of which properties of iterated multiple 
	revision should be deemed desirable, and of how we may apply the 
	principles underlying the AGM theory of belief revision in the 
	context of default reasoning. 

\section{Acknowledgements}	

	This work is supported by the DFG project ``Computationale Dialektik''
	within the DFG research group ``Kommunikatives Verstehen''.
	Much of this paper was written while the author was a 
	researcher at the Max-Planck-Institute for Computer Science in
	Saarbr\"ucken, Germany. The author would like to thank Emil 
	Weydert, Michael Freund,
	Hans Rott, Leon van der Torre and the anonymous referees for helpful 
	comments and suggestions.
	
\bibliography{nmr}
\bibliographystyle{aaai}

\end{document}